# AI-based Arabic Language and Speech Tutor

Sicong Shao, Saleem Alharir,
Salim Hariri
NSF Center for Cloud and Autonomic
Computing
University of Arizona,
Tucson, Arizona AZ 85721 USA

Pratik Satam
Department of Systems and Industrial
Engineering
University of Arizona,
Tucson, Arizona AZ 85721 USA

Sonia Shiri, Abdessamad Mbarki
Arabic Special Programs
School of Middle Eastern and North
African Studies
University of Arizona,
Tucson, Arizona AZ 85721 USA

*Abstract*— In the past decade, we have observed a growing interest in using technologies such as artificial intelligence (AI), machine learning, and chatbots to provide assistance to language learners, especially in second language learning. By using AI and natural language processing (NLP) and chatbots, we can create an intelligent self-learning environment that goes beyond multiple-choice questions and/or fill in the blank exercises. In addition, NLP allows for learning to be adaptive in that it offers more than an indication that an error has occurred. It also provides a description of the error, uses linguistic analysis to isolate the source of the error, and then suggests additional drills to achieve optimal individualized learning outcomes. In this paper, we present our approach for developing an Artificial Intelligence-based Arabic Language and Speech Tutor (AI-ALST) for teaching the Moroccan Arabic dialect. The AI-ALST system is an intelligent tutor that provides analysis and assessment of students learning the Moroccan dialect at University of Arizona (UA). The AI-ALST provides a self-learned environment to practice each lesson for pronunciation training. In this paper, we present our initial experimental evaluation of the AI-ALST that is based on MFCC (Mel frequency cepstrum coefficient) feature extraction, bidirectional LSTM (Long Short-Term Memory), attention mechanism, and a cost-based strategy for dealing with class-imbalance learning. We evaluated our tutor on the word pronunciation of lesson 1 of the Moroccan Arabic dialect class. The experimental results show that the AI-ALST can effectively and successfully detect pronunciation errors and evaluate its performance by using $F_1$ – score, accuracy, precision, and recall.

## I. INTRODUCTION

Recently, there has been a strong interest in technology-driven language learning using interactive websites, artificial intelligence, synchronous chat, and virtual environments to assist language learners intelligently. These efforts used artificial intelligence agents such as chatbots that can have a significant impact, but their efficacies in teaching second languages have not been well explored due to the fact that the technology is still under development and has not been widely applied [1, 2]. Moreover, automated language teaching technology is still underexplored for low-resource languages (e.g., Arabic dialects), which lack enough electronic resources for speech and language processing. Therefore, it is beneficial to explore how to assist second language learning through AI-driven chatbots, especially in budget-constrained environments and due to the lack of sufficient low-resource language datasets and knowledgeable human tutors [3, 22].

In general, an intelligent tutoring system can be described in terms of three modules: the domain knowledge, the student model, and the teaching module [4]. In this environment, students can learn at their own pace, and the student model can help us develop individualized knowledge-based instruction according to domain knowledge learning and teaching. The teaching model guides the process of learning and teaching [5]. The student model maintains the learner history and provides the capability to update the student learner model and hence allowing the system to adapt to student needs and alter the instructional process accordingly. Developing speaking in foreign or second languagesis generally better served by conversational interaction [12] and intercultural communication [23] than through the study of grammar. Furthermore, practical speaking exercises in second language learning are better conducted in a personalized manner via a tutor, which represents a costly teaching mode due to the lack of available and knowledgeable tutors and the intensity of the process. These problems can be overcome if we offer students an intelligent self-study environment that can be accessed 24/7 using any Internet-ready device (mobile or stationary).

Computer-assisted pronunciation training (CAPT), as an essential component of the language learning system, assists learners in improving their speaking skills by providing feedback for pronunciation error detection [6]. Machine learning is one of the most promising ways to implement CAPT since it automatically and efficiently provides the learner with personalized feedback for correcting detected errors [7]. This paper focuses on machine learning-based CAPT with an intelligent tutor for Arabic pronunciation detection. We present our approach to develop an AI-ALST for teaching the Moroccan Arabic dialect, which is an Arabic dialect unexplored in previous machine learning-based CAPT research. The AI-ALST system provides intelligent tutoring that analyzes and assesses students learning the Moroccan dialect at the UA School of Middle Eastern and North African Studies. The CAPT component of AI-ALST is constructed based on MFCC, bidirectional LSTM, attention mechanism, and cost-based class-imbalance learning strategy.

The organization of the remaining sections of the paper is as follows. In Section II, we present the related work for language learning systems and chatbot technologies for intelligent tutor. Section III reviews the Arabic learning programs at The University of Arizona. In Section IV, we describe the architecture of the AI-ALST system being developed at UA. In Section V, we present our preliminary experimental results and evaluation of the system when applied to Lesson 1 of the Arabic Moroccan dialect class. In section VI, we present summary of the research results and future research plans.

## II. RELATED WORK

The development of advanced language learning systems in general and second language in particular have followed two approaches: Corpus-based Learning and AI-based

Tutoring systems. In what follows, we briefly discuss each approach.

The number of open source corpora that can be used by anyone with Internet-ready device has increased significantly over the past couple of decades for many languages. Language teaching is one of the areas that has benefited from progress in corpus development. However, the number of corpus-based teaching in the Arabic language is very limited compared to other languages [8]. We have seen major progress in the availability of Arabic corpus resources in the last few decades, but they are mainly used by researchers and less for teaching Arabic as a foreign language. The main developers of Arabic language corpora include European Language Resources Association (ELRA), Linguistic Data Consortium (LDC) and Research and Development International (RDI) [8]. ELRA developed Al-Hayat Arabic Corpus (18 million words), Al-Nahar text corpus (24 million words), AFP corpus, and NEMLAR written corpus (500,000 words). The LDC developed Arabic Gigaword (400 million words) and Arabic Newswire corpus (80 million words). RDI developed Arabic morphological analysis, PoS tagging, Arabic discretization, and phonetic transcription, and lexical semantic labeling. This list is not comprehensive, but it shows that we need more open-source Arabic corpora and the tools to use them for research and teaching Arabic as a second language. One corpus that focuses on teaching Arabic as a second language is the Arabic Learner Corpus (ALC). ALC is a valuable resource for conducting research on Arabic teaching and learning as well as Arabic NLP. It includes 282,732 words and 1,585 materials (written and spoken) produced by 942 students from 67 nationalities and 66 different backgrounds. ALC is one of the largest learner corpora for Arabic that comprises data from both native Arabic speakers and non-native Arabic speakers, and the first Arabic learner corpus for Arabic as a Second Language (ASL1) [9].

Intelligent-based tutoring systems that use intelligent agents (chatbots) are especially important to foreign language learners due to their ability to provide self-study with error-contingent feedback, adaptable learning processes, and friendly environments [1]. Chatbots are software systems that automatically communicate with users via text or speech format [11, 12, 24]. Eliza was the first chatbot, which was developed by the Massachusetts Institute of Technology (MIT) in 1966 [10]. Since then, chatbots have been used in various fields such as healthcare, business, education, and banking [13]. Recently, due to COVID-19 pandemic, we have seen a rapid acceptance of chatbots usage. The research of Arabic dialect chatbot is still limited. BOTTA, the first Arabic dialect chatbot, aims to create friendly conversations through the Egyptian Arabic dialect [27]. Nabiha is another Arabic dialect chatbot that can support conversations using the Saudi Arabic dialect [26]. Chatbots architecture consists of three modules: Web-interface, conversation module and knowledge base. The web-interface can use text, voice, or even instant messaging app (e.g., WhatsApp). The conversation module uses AI and NLP to understand the semantics and intentions of the conversation [25]. The knowledge base stores the rules and content that govern the chatbot operations based on three types [13]: rule-based, retrieval-based, and generative-based. The main challenge of chatbot technology facing is creating the feeling that the user is talking to a real-person. AI-powered chatbots such as Facebook messenger bots, Alexa, Microsoft Cortana, and Siri [3] have made good progress, but they are mainly in English. The development of Arabic language chatbots faces many challenges due to the lack of Arabic corpora that are suitable for developing Arabic language chatbots. In addition, the Arabic language has rich morphology, orthographic variations, ambiguity, and multiple dialects [13].

### III. THE COURSE: MOROCCAN ARABIC DIALECT

As proof of concept, we use materials from the learning modules of two Moroccan Arabic multimedia eTextbooks to experiment with the AI-ALST and evaluate its performance. These eTextbooks were selected for this experiment because they already integrated technology tools, albeit of the conventional type. The courses aim at introducing learners to the variety of Arabic spoken in Morocco and the culture of the country. Their goal is to prepare students for interaction in various types of informal and formal situations. Both courses are organized around seven specific modules based on a 360 needs assessment survey of Arabic Language Flagship Program learners, directors, overseas community members and instructors. The modules prepare learners to communicate in a variety of contexts identified as important while living and studying in Morocco.[1] Approximately two to three semesters of study are typically needed to get students to the Intermediate Low level of Arabic. Therefore, time limitations make it challenging for Flagship students to gain sufficient proficiency in Moroccan Darija using conventional approaches and materials, hence the introduction of the simple technology-enhanced content in these courses.

Participants learn how to understand, discuss, and reflect on the themes of the courses using a variety of technology-based materials. The courses were intended to be used in the blended/hybrid format. This means that learning materials and assignments that are available online are combined with instructor-led interactive sessions, whether online or in person. Homework is assigned in advance of the class following a flipped-classroom model. Students are expected to be fully prepared to participate in activating the language through role play and discussions in class, whether online or in person.

Level I is for beginning Moroccan Darija with the final module dedicated to providing essentials about work environments. It is intended for students with little or no prior Moroccan Darija. Level II is for students with prior Moroccan Arabic either through formal instruction or study abroad. The Level I course contains seven modules, and each module consists of three lessons. The first session of the course consists of an introduction to Moroccan Arabic (Darija) and the relationship between Darija and Modern Standard Arabic and provides an overview of the course syllabus and the technical requirements. The remaining lessons are each divided into three main sections (vocabulary and structures, listening comprehension, and speaking assessment). At the end of the module, there is a module assessment to evaluate what learners achieved.

The Level II course focuses on the same themes that were identified during the needs assessment that was conducted in preparation of the course design. The seven themes cover various contexts that Arabic Flagship students will encounter while in Morocco and the language and culture needed to operate in those contexts. The level two course was built on the materials covered in the basic level course and provided depth regarding those seven modules. Through a combination of technology-mediated materials, learners are able to use Moroccan Arabic at a more complex and extended level of discourse and navigate the cultural norms associated with these contexts. Each one of these modules contains two lessons and each lesson comes with its own authentic video content and accompanying activities.

[1] The courses were prepared in collaboration with the Language Flagship Technology Innovation Center at the University of Hawai`i at Manoa with funding from the National Security Education Program.

The AI-ALST system is using the Moroccan Arabic Darija as a use case to demonstrate how to use the system to provide intelligent self-learning tutor to students.

## IV. AI-ALST ARCHITECTURE

The overview of AI-ALST architecture is shown in Figure 1. First, the learner selects the material from vocabulary, sentences, or conversations that are associated with each lesson in the class. Then, students' pronunciations of each word will be sent in real-time to the pronunciation tutoring engine, which outputs the pronunciation analysis result. After that, the feedback that will be provided to the student indicates if the pronunciation was correct, identifies the parts that contributed to the mispronunciation, and the recommended next step. In this paper, we report our preliminary research results of applying AI-ALST to lesson 1. Future development will provide intelligent feedbacks on the source of the error and recommends the appropriate exercises to fix the individual wrong pronunciation associated with each student.

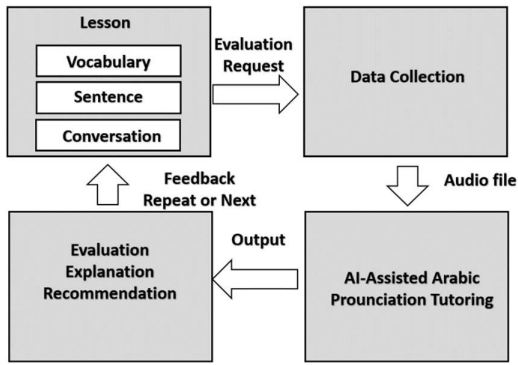

Fig. 1. The overview of AI-ALST architecture.

### A. Overview of Deep Learning-based Pronunciation Tutoring

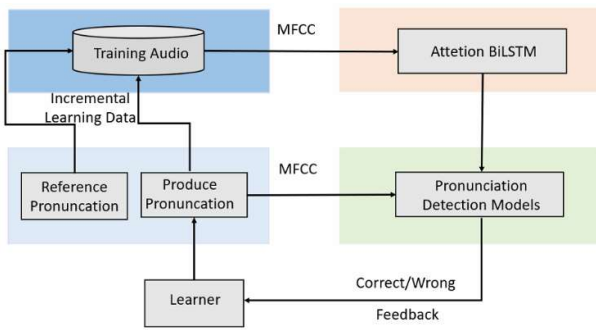

Fig. 2. The architecture of the pronunciation tutoring phase of AI intelligent tutor.

The architecture of the pronunciation tutoring part of the AI-ALST system is shown in Figure 2. First, the training data set consists of recorded audios of pronunciations that are converted to Mel frequency cepstrum coefficient (MFCC) features. We train attention BiLSTM to obtain a pronunciation detection model for each word. After finishing the training stage, the models are deployed to detect pronunciation errors. A learner who wants to study the correct pronunciation of a word can first learn the correct pronunciation from the reference pronunciation provided by a native speaker and then the student attempts to pronounce this word to be as close as possible to the reference pronunciation. The pronunciation detection model identifies whether the word pronunciation being analyzed is correct or wrong and promptly gives the student feedback. Furthermore, the pronunciation detection model supports incremental learning by using new data from experts and learners that will significantly improve to improve AI-ALST system performance.

### B. MFCC Feature Extraction

MFCC features are one of the most successful feature extraction approaches for automatic speech recognition [14]. It can capture a broad frequency range and incorporate perceptual weighting for frequency bands. Therefore, we use MFCC features to represent the word audio of pronunciation. Figure 3 shows the main steps of MFCC feature extraction, which are described as follows: (1) The pronunciation audio is pre-emphasized by a filter to flatten the word audio spectrally. (2) The audio signal is separated into a number of frames, where each frame is processed by a hamming window to minimize audio signal discontinuities and reduce the edge effect. (3) Discrete Fourier transform (DFT) is applied to each frame, where fast Fourier transform is utilized to implement DFT for improving efficiency. (4) Mel spectrum is calculated using the Fourier transformed audio signal through Mel filter bank consisting of a number of band-pass filters. (5) Discrete cosine transform (DCT) is used on the transformed Mel-frequency coefficients to generate cepstral coefficients. The Mel spectrum is represented on a log scale before DCT is computed. In this study, the first 20 dimensions of the MFCC features were extracted.

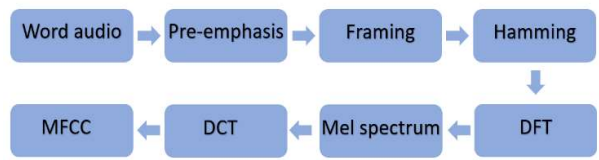

Fig. 3. The steps of MFCC feature extraction.

### C. Attention BiLSTM-based Pronunciation Detection Model

The attention BiLSTM-based pronunciation detection model is proposed for the pronunciation detection task of the intelligent tutor. As shown in Figure 4, the detection model consists of several parts: an MFCC feature extraction layer, a BiLSTM layer combining forward and backward LSTM, an attention layer, and an output layer with a cost-based strategy for dealing with class-imbalanced audio data. The details of each part of the model are illustrated in the following paragraphs.

Recurrent neural network (RNN) is a type of deep learning algorithm used for modeling sequential data [15]. Simple RNNs are able to extract short-term dependency, but they cannot learn long-term dependencies well because of the vanishing and exploding gradient problem. LSTM alleviates the vanishing/exploding gradient by including an additional recurrent loop that ensures past information is retained unless the forget gate removes it [16]. As shown in Figure 5, an LSTM consists of cells whose outputs are influenced by past memory content. Cells in the LSTM chain have a common cell state, maintaining long-term dependency. The input gate $i_t$ and forget gate $f_t$ control the flow of information, allowing the network to update the current state $C_t$ or forget the previous state $C_{t-1}$. The output gate ($o_t$) controls the output of the hidden state (each cell), which enables the cell to compute its output based on the updated cell state. The following formulas describe an LSTM cell architecture that is used to process the pronunciation audio input $x_t$ at the current timestep:

$$i_t = \sigma(W_i x_t + W_i h_{t-1} + b_i) \quad (1)$$

$$f_t = \sigma(W_f x_t + W_f h_{t-1} + b_f) \quad (2)$$

$$g_t = \tanh(W_c x_t + W_c h_{t-1} + b_c) \quad (3)$$

$$C_t = f_t \odot C_{t-1} + i_t \odot g_t \quad (4)$$

$$o_t = \sigma(W_o x_t + W_o h_{t-1} + b_o) \quad (5)$$

$$h_t = o_t \odot \tanh(C_t) \quad (6)$$

where $W_i, W_f, W_o, W_c$ represent the weight parameters, and $b_i, b_f, b_o, b_c$ are the bias parameters, and $\sigma$ is sigmoid function. The free parameters of LSTM can be learned by backpropagation through time.

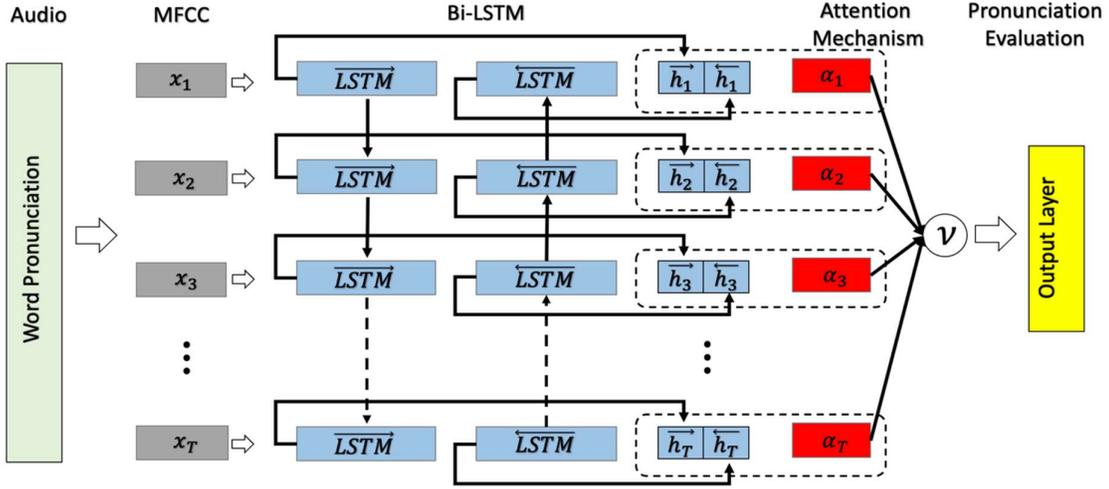

Fig. 4. The pipeline of the attention BiLSTM-based pronunciation detection model structure, which includes MFCC feature extraction, BiLSTM layer, attention layer, and output layer with cost-based strategy for class-imbalance learning.

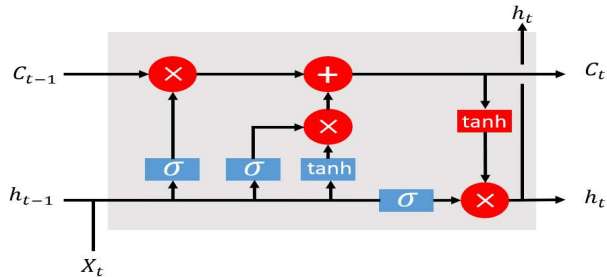

Fig. 5. The architecture of an LSTM cell.

The bidirectional LSTM (BiLSTM) model is an extension of the LSTM models where the input data is processed using two LSTMs. In many research fields, including sequence classification, BiLSTM shows superior performance compared with LSTM [17]. As shown in Figure 4, the BiLSTM applies the sequential data into the forward layer and also provides the reversed copy of the sequential data into the backward layer. Hence, the hidden states can utilize past and future information to generate feature representation. The hidden states of forward direction and backward direction are concatenated as the outputs of BiLSTM given by:

$$\overrightarrow{h_t} = \overrightarrow{LSTM}(x_t), t \in [1, T] \quad (7)$$

$$\overleftarrow{h_t} = \overleftarrow{LSTM}(x_t), t \in [T, 1] \quad (8)$$

$$h'_t = [\overrightarrow{h_t}, \overleftarrow{h_t}], t \in [1, T] \quad (9)$$

therefore, in our study, with BiLSTM hidden states $h'_t$, the BiLSTM model can effectively extract and incorporate contextual information from pronunciation audio and leads to improving the learning of long dependency and thus, consequently, will benefit the word pronunciation correction tasks.

The attention mechanism aims to give varying weights to each timestep of sequential data and selects information relevant to the task by learning context vectors in conjunction with other parameters during the training process [18]. Attention mechanisms can improve the learning process's effectiveness by exploring the data's inherent features.

In the pronunciation detection task, we conjecture that not all audio frames have an equal contribution to the representation for the pronunciation detection of audio. However, BiLSTM model still has limitations in extracting discriminative information in audio frames to evaluate the pronunciation. Therefore, to further improve the detection performance, we employ the attention mechanism in our BiLSTM deep learning model to reward the frames that are hints for detecting the pronunciation effectively. Concretely, the hidden states $h'_t, t \in [1, T]$ generated by Bi-LSTM is used as the input of a simple multilayer perceptron to obtain a hidden representation $u_t$ for $h'_t$. After that, we calculate the similarity of $u_t$ with context vector $u_w$ to evaluate the importance of the audio frame, which can be jointly learned during the model training. Then, the normalized importance weight $\alpha_t$ indicating the importance of the timestep $t$ of the pronunciation audio input is obtained using the softmax function. Therefore, the process of each timestep can be formulated as follows:

$$u_t = \tanh(W_w h'_t + b_w) \quad (10)$$

$$\alpha_t = \frac{exp(u_t^T u_w)}{\sum_t exp(u_t^T u_w)} \quad (11)$$

$$v = \sum_t \alpha_t h'_t \quad (12)$$

where vector $v$ is the output of the attention layer as the high-level representation of an audio of pronunciation, and $W_w$ and $b_w$ are free parameters learned during the training.

For giving a final detection result of an audio of pronunciation, a fully connected layer is used on top of the vector $v$ generated by the attention layer to get a prediction:

$$p = \sigma(W_z v + b_z) \quad (13)$$

where $p$ is prediction probability, $\sigma$ is a sigmoid function, and $W_z$ and $b_z$ are trainable parameters. After training with the data given two-class label, the model predicts $Y = 1$ indicating the pronunciation is wrong when $p \geq 0.5$ and $Y = 0$ indicating the pronunciation is correct when $p < 0.5$. The binary cross-entropy is leveraged to calculate the loss during the learning, and Adam [21] is used as the optimizer.

Class-imbalance data occurs in our two-class classification problem. The data of the minority class is significantly less than the data of the majority class. The pronunciation error classification performance may be suboptimal if we do not handle the class imbalance problem. To deal with this problem, a cost-based strategy [19] is employed because it can be seamlessly integrated into our attention-based BiLSTM model. Specifically, the weight is attributed to each training sample during the training, and larger weights are applied to the class with smaller amounts, which are the training samples with higher cost if they are classified incorrectly. In contrast, lower weights are assigned to the class with larger amounts, which are the training samples with lower costs if they are wrongly classified.

## V. EXPERIMENTS AND RESULTS

In this section, we conduct experiments on the collected dataset and present our results. To evaluate the performance of the proposed approach. we apply our approach to a dataset for Moroccan Arabic language pronunciation error detection.

### A. Experimental Settings

The dataset is a corpus of 17 common words from a Moroccan Arabic Darija class. The words are pronounced by 12 participants. There are 3,851 audios in the dataset and each audio recorded the pronunciation of a word from a participant. Among those audios, 2,961 audios are correct pronunciations, and 890 audios are mispronunciations. We treat each audio as a classification instance, and all the audios are processed by padding method [7] to obtain equal duration.

Since we independently train a binary classification model for each word in the lesson, we launch 17 experiments where each experiment evaluates the performance of the approach on a particular word. Hence, in each experiment, we perform random sampling on the audios of correct pronunciations of that word to extract 50% of audios as correct pronunciation category data in the training set and the remaining 50% of audios as the correct pronunciation category data in the test set. Also, random sampling is performed on the audios of mispronunciations of that word to obtain 50% audios as mispronunciation category data in the training set and the remaining 50% of audios as the mispronunciation category data in the test set. For testing a word, due to its limited mispronounced audios in the dataset, the data is not enough to represent the mispronunciation category. Therefore, in order to enhance the diversity and get a better representative of the mispronunciation category data during the training, we also include the correct pronunciation audios from the remaining 16 words as the mispronunciation category when we perform training for that word. We should note that any different Moroccan Arabic words with different pronunciations can also be added to mispronunciation category data for training, although we have not added more words due to the limited data we collected. After adding the remaining words to the training data, the correct pronunciation category audio samples are much less than the mispronunciation category audio samples in the training data in each word pronunciation classification. However, please note that the test set for each word it only includes the correct pronunciation and mispronunciation of that word and does not contain audios from other words. In the current experiment, we faced the class imbalance problem in both training and test data sets for each word Therefore, the cost-based strategy is employed for training the deep learning models to address the class imbalance problem.

### B. Evaluation Metric

The common accuracy metric cannot well assess class imbalance problem [20]. Therefore, we use more suitable metrics for performance evaluation such as precision, recall, and $F_1$-score which is a combined metric of precision and recall. In our experimental result evaluation, we measure the pronunciation classification results on all the test sets according to four cases: (1) True Positive (TP), meaning both the detection model and the label report mispronunciation; (2) True Negative (TN), meaning both the detection model and the label report correct pronunciation; (3) False Positive (FP), meaning the detection model predicts mispronunciation, but the label report correct pronunciation; and (4) False Negative (FN), meaning the detection model predicts correct pronunciation, but the label reports mispronunciation. We calculated the evaluation metrics using the counts from each case, which are given by:

$$Precision = \frac{TP}{TP + FP} \quad (14)$$

$$Recall = \frac{TP}{TP + FN} \quad (15)$$

$$Accuracy = \frac{TP + TN}{TP + TN + FP + FN} \quad (16)$$

$$F_1 - score = \frac{2 \times Precision \times Recall}{Precision + Recall} \quad (17)$$

TABLE I AVERAGE F1 (%) AND ACCURACY (%) WITH STANDARD DEVIATION (%) PER METHOD OVER FIVE RANDOM SEEDS

| Word | | $F_1$-score | | Accuracy | |
|---|---|---|---|---|---|
| Moroccan | English | BiLSTM | Attention BiLSTM | BiLSTM | Attention BiLSTM |
| السّلام عليكم | Greetings! (generic greeting) | 76.3 ± 4.9 | **85.5 ± 3.3** | 93.3 ± 1.5 | **96.2 ± 0.9** |
| وعليكم السّلام | Greetings to you! (response) (a generic greeting) | 56.6 ± 15.9 | **89.0 ± 3.5** | 82.5 ± 3.8 | **94.4 ± 1.8** |
| كيداير؟ | How are you (masc.)? | 67.0 ± 2.0 | **87.9 ± 12.0** | 87.1 ± 1.1 | **95.2 ± 4.3** |
| كيدايرة؟ | How are you (fem.)? | 76.0 ± 7.2 | **86.1 ± 7.5** | 90.9 ± 2.3 | **94.1 ± 2.5** |
| آش خبارك؟ | How are you? (type-1) | 86.6 ± 1.8 | **91.4 ± 1.4** | 90.1 ± 1.3 | **94.1 ± 1.1** |
| لابس عليك؟ | How are you? (type-2) | **95.2 ± 1.6** | 93.8 ± 1.4 | **98.0 ± 0.7** | 97.3 ± 0.6 |
| مزيان | Good. | **94.0 ± 1.8** | 93.2 ± 2.9 | **97.1 ± 0.9** | 96.6 ± 1.5 |
| شويّة | So-so; OK. | 67.0 ± 7.3 | **77.2 ± 15.2** | 88.9 ± 2.1 | **91.5 ± 4.5** |
| لاباس الحمد لله، وأنت؟ | I am good, and yourself? | 91.9 ± 3.3 | **96.8 ± 2.1** | 93.6 ± 2.6 | **97.6 ± 1.6** |
| كل شي مزيان؟ | Is everything good? | **87.4 ± 4.0** | 85.5 ± 13.6 | 91.4 ± 3.3 | **91.7 ± 6.1** |
| كل شي مزيان | Everything is fine. | 66.1 ± 8.0 | **84.8 ± 7.9** | 85.7 ± 2.6 | **92.6 ± 2.9** |
| بارك الله فيك | Thank you. | 85.2 ± 2.8 | **89.1 ± 4.1** | 92.4 ± 1.6 | **94.8 ± 1.9** |
| الله يبارك فيك | Thank you (response). | 90.2 ± 2.7 | **95.6 ± 0.8** | 96.0 ± 1.0 | **98.2 ± 0.3** |
| كل شي بخير، الحمد لله | Everything is good. | 85.8 ± 6.3 | **92.8 ± 1.7** | 93.0 ± 2.8 | **96.6 ± 0.8** |
| سلّم على | Say hi to... (masc.) | **87.5 ± 4.5** | 86.6 ± 4.5 | **94.1 ± 2.1** | 93.1 ± 2.7 |
| سلّمي على | Say hi to... (fem.) | **83.4 ± 2.4** | 77.7 ± 4.9 | **90.5 ± 1.6** | 85.9 ± 4.2 |
| سلّمي على الوالدة | Say hi to your mother. | 82.5 ± 3.0 | **84.4 ± 3.8** | 90.5 ± 1.7 | **91.8 ± 2.1** |

TABLE II AVERAGE PRECISION (%) AND RECALL (%) WITH STANDARD DEVIATION (%) PER METHOD OVER FIVE RANDOM SEEDS

| Word | | Precision | | Recall | |
|---|---|---|---|---|---|
| Moroccan | English | BiLSTM | Attention BiLSTM | BiLSTM | Attention BiLSTM |
| السّلام عليكم | Greetings! (generic greeting) | 72.8 ± 7.0 | **88.1 ± 5.2** | 80.7 ± 5.0 | **83.2 ± 4.3** |
| وعليكم السّلام | Greetings to you! (response) (a generic greeting) | 66.4 ± 6.3 | **84.5 ± 4.2** | 52.9 ± 22.1 | **94.2 ± 4.3** |
| كيداير؟ | How are you (masc.)? | 88.7 ± 7.3 | **96.8 ± 2.8** | 54.0 ± 1.3 | **83.3 ± 20.1** |
| كيدايرة؟ | How are you (fem.)? | **96.1 ± 5.6** | 91.6 ± 3.1 | 63.7 ± 9.8 | **83.0 ± 14.4** |
| آش خبارك؟ | How are you? (type-1) | 80.2 ± 1.5 | **90.4 ± 3.6** | **94.2 ± 3.3** | 92.5 ± 1.7 |
| لابس عليك؟ | How are you? (type-2) | **98.5 ± 3.0** | 95.3 ± 2.9 | 92.3 ± 3.4 | **92.3 ± 0.0** |
| مزيان | Good. | **96.2 ± 3.3** | 91.7 ± 4.8 | 91.9 ± 1.5 | **94.8 ± 1.8** |
| شويّة | So-so; OK. | **79.2 ± 7.0** | 77.1 ± 12.0 | 58.6 ± 8.6 | **79.3 ± 20.8** |
| لاباس الحمد لله، وأنت؟ | I am good, and yourself? | 86.4 ± 5.0 | **94.5 ± 3.1** | 98.3 ± 2.0 | **99.2 ± 1.7** |
| كل شي مزيان؟ | Is everything good? | 91.0 ± 9.2 | **93.4 ± 4.2** | **85.0 ± 5.7** | 82.5 ± 20.7 |
| كل شي مزيان | Everything is fine. | 82.7 ± 3.9 | **87.2 ± 8.8** | 55.8 ± 10.7 | **85.8 ± 15.9** |
| بارك الله فيك | Thank you. | 84.6 ± 5.1 | **93.5 ± 4.7** | **86.2 ± 3.9** | 85.4 ± 6.6 |
| الله يبارك فيك | Thank you (response). | 95.8 ± 2.5 | **99.2 ± 1.6** | 85.4 ± 5.1 | **92.3 ± 0.0** |
| كل شي بخير، الحمد لله | Everything is good. | 82.5 ± 5.0 | **94.9 ± 3.0** | 90.0 ± 10.1 | **90.8 ± 3.1** |
| سلّم على | Say hi to... (masc.) | **84.6 ± 5.6** | 78.1 ± 7.1 | 90.8 ± 5.2 | **97.7 ± 1.9** |
| سلّمي على | Say hi to... (fem.) | **73.7 ± 3.8** | 66.1 ± 10.0 | **96.2 ± 0.0** | 96.2 ± 4.2 |
| سلّمي على الوالدة | Say hi to your mother. | 71.4 ± 3.3 | **74.4 ± 3.9** | **97.7 ± 3.1** | 97.7 ± 4.6 |

As our dataset is class imbalanced and the main goal is to detect mispronunciation, precision and recall are employed for model evaluation. The precision is used to measure whether the model can accurately detect mispronunciation. The recall refers to the capability to correctly detect mispronunciations among all the mispronunciations. The $F_1 - score$ provides a balanced overview of precision and recall.

*C. Training Methodology*

To recognize a pronounced word, we extract features from the audio waveform using MFCC. The first 20 MFCC features are used in this paper. For each pronounced word, we train the attention-based BiLSTM model using same hyper-parameters to address the robustness of our approach. Concretely, we set the LSTM layer dimension to 128. Therefore, the BiLSTM combining forward and backward LSTM has 256 dimensions. We train the models with 64 mini-batch size, using Adam optimizer and binary cross entropy loss. Our cost-based strategy addresses class imbalance data, i.e., training audio samples from the different classes are given with different weights. To validate our pronunciation error detection approach, attention BiLSTM model with MFCC features is compared with BiLSTM model with same MFCC features. For the experiment of each word, we train the deep learning models with five different random seeds and report the mean performance with standard deviation.

*D. Results*

The results are shown in Table I and Table II. In most cases, the attention BiLSTM shows good pronunciation detection performance, validating our approach's effectiveness. What's more, the attention BiLSTM-based models perform much better than the BiLSTM-based models for the overall performance of the four evaluation metrics. For the $F_1$-score and recall, the results of 12 words of attention BiLSTM are better than BiLSTM. For accuracy, attention BiLSTM performs better than BiLSTM on 13 words. Besides, attention BiLSTM attains better precision on 11 words.

The BiLSTM has achieved some success, however, there are still six words whose $F_1$-score are lower than 80% and five words whose precision and recall are lower than 80%, which is unsatisfactory. While only two words' $F_1$-score and four words' precision and one word's recall are lower than 80% for attention BiLSTM, which beats BiLSTM. We also notice that attention BiLSTM outperforms the counterpart by large margin for the overall measurements of the words including "Greetings to you!", "How are you (masc.)?", "How are you (fem.)?", "So-so; OK", and "Everything is fine".

## VI. CONCLUSION

In this paper, we presented a novel architecture to implement an Artificial Intelligence-based Arabic Language and Speech Tutor (AI-ALST) for teaching the Moroccan Arabic dialect. The AI-ALST system provides analysis and assessment of students learning the Moroccan dialect at the University of Arizona (UA) School of Middle Eastern and North African Studies. The AI-ALST provides a self-learned environment to practice each lesson for pronunciation training. The AI-ALST implementation is based on MFCC feature extraction, bidirectional LSTM, attention mechanism, and a cost-based strategy for dealing with class-imbalance learning. We evaluated our tutor on the words pronunciation of lesson 1 of the Moroccan Arabic dialect class. The experimental results show that the AI-ALST can effectively and successfully detect pronunciation errors and evaluate its performance by using $F_1$-score, accuracy, precision, and recall

The AI-ALST system can be applied to any language (e.g., other Arabic dialects) learning as long as there is enough training data. In future work, we will apply the AI-ALST system to provide self-learning for the remaining lessons in the courses and learning modules. Another future research issue is to add to the AI-ALST system the capability to adaptively adjust the tutoring exercises and recommendations to match the current cognitive level of the learner. Moreover, we also attempt to improve feedback by providing explainable diagnoses for detected pronunciation errors and suggest other words to help the learning in fixing the detected wrong pronunciation. Finally, we plan to use explainable machine learning with the intelligent tutor to study the impact of the first language in the accent on the learning process of Moroccan Arabic dialects.

ACKNOWLEDGMENT

This work is partly supported by National Science Foundation (NSF) research projects NSF-1624668 and NSF-1849113, (NSF) DUE-1303362 (Scholarship-for-Service), and Department of Energy/National Nuclear Security Administration under Award Number(s) DE-NA0003946.